%% file: paper.tex
\documentclass{article}

\usepackage{arxiv}

\usepackage[utf8]{inputenc} 
\usepackage[T1]{fontenc}    
\usepackage{hyperref}       
\usepackage{url}            
\usepackage{boldline,booktabs}       
\usepackage{amsfonts,amsmath}       
\usepackage{nicefrac}       
\usepackage{microtype}      
\usepackage{lipsum}
\usepackage{natbib}
\usepackage{enumitem}
\usepackage[dvipsnames]{xcolor}
\usepackage{varwidth}
\usepackage{tikz}
\usepackage{tkz-euclide}
\usetikzlibrary{hobby,arrows,backgrounds,shapes,fit,positioning,calc,shadows,matrix,intersections,fadings,shapes.arrows,decorations.pathreplacing}
\usetkzobj{all}
\usepackage{pgfplots}
\usepackage{algorithm2e}

\newcommand{\R}{\mathbb{R}}
\newcommand{\V}{\mathbf{v}}
\newcommand{\W}{\mathbf{w}}
\DeclareMathOperator*{\kNN}{kNN}
\DeclareMathOperator*{\mkNN}{mkNN}
\DeclareMathOperator*{\IC}{IC}
\DeclareMathOperator*{\CS}{CS}
\newcommand{\C}{\mathcal{C}}
\renewcommand{\S}{\mathcal{S}}

\tikzfading[name=arrowfading, top color=transparent!0, bottom color=transparent!95]
\tikzset{arrowfill/.style={top color=CornflowerBlue!20, bottom color=Cerulean, general shadow={fill=black, shadow yshift=-0.4ex, path fading=arrowfading}}}
\tikzset{arrowstyle/.style={draw=FireBrick,arrowfill, single arrow,minimum height=#1, single arrow,
single arrow head extend=.2cm,}}
\definecolor{FireBrick}{rgb}{0.7, 0.13, 0.13}

\newcommand{\tikzfancyarrow}[2][2cm]{\tikz[baseline=-0.5ex]\node [arrowstyle=#1] {#2};}

\tikzset{
    position/.style args={#1:#2 from #3}{
        at=(#3.#1), anchor=#1+180, shift=(#1:#2)
    }
    }

\title{A flexible outlier detector based on a topology  given by graph communities}

\author{
  O. Ramos Terrades, A. Berenguel, D. Gil\thanks{DGil is a Serra Hunter Fellow} \\
  Computer Vision Center - Dept. of Computer Science\\
  Universitat Aut\`{o}noma de Barcelona\\
  Cerdanyola del Vallés, 08193 \\
  \texttt{{oriolrt,aberenguel,debora}@cvc.uab.cat} \\
}

\begin{document}
\maketitle

\begin{abstract}




Outlier, or anomaly, detection is essential for optimal performance of machine learning methods  and statistical predictive models. It is not  just a technical step in a data cleaning process but a  key topic in many fields  such as fraudulent document detection,   in medical applications and assisted diagnosis systems or detecting security threats. In contrast to population-based methods, neighborhood based local approaches are simple flexible methods that have the potential to perform well in small sample size unbalanced problems. However, a main concern of local approaches is the impact that the computation of each sample neighborhood has on the method performance. Most approaches use a distance in the feature space to define a single neighborhood that requires careful selection of several parameters.

This work presents a local approach based on a local measure of the heterogeneity of sample labels in the feature space considered as a topological manifold. Topology is computed using the communities of a weighted graph codifying mutual nearest neighbors in the feature space. This way, we provide with a set of multiple neighborhoods able to describe the structure of complex spaces without parameter fine tuning. The extensive experiments on real-world  data sets show that our approach overall outperforms, both, local and global strategies in multi and single view settings.

\end{abstract}


\section{Introduction}

Outlier, or anomaly, detection is a major issue in many learning-based algorithms since the  presence of outlier data on training data might affect the proper estimation of model parameters. Moreover, outlier detection becomes harder when data changes along time, since it is unclear how to distinguish proper data, coming from a new arising class, from corrupted, or outlier, data. Outlier detection is not just a technical step in a data cleaning process. It is, by itself, a key topic in many fields  such as fraudulent document detection, in identity documents or passports, insurances claims and health care fraud; medical applications and assisted diagnosis systems; detection of security threats, etc.  In all these tasks, an accurate outlier detection will impact in the economic balance of any company or to properly detect any security threat, for instance. 

The outlier concept is  fuzzy, it  depends on each task and thus each detection method provides its own  definition. For instance, \cite{chandola2009anomaly} define an outlier as a sample that ``does not conform to expected behavior'' and they classify them as {\em point outliers}, {\em contextual outliers} and {\em collective outliers}. Point outliers are samples with abnormal feature values not expected for any of the classes and they correspond to isolates samples either not belonging to any cluster or not following the population distribution. Contextual outliers are samples which are labelled differently from their neighboring samples, which define their context. Collective outliers are small group of samples sharing unusual features that are clustered together. In \cite{Zhao2018} outliers are split into {\em attribute outliers} and {\em class outliers} in the context of multimodal representations. In such representations there are two or more feature spaces (called {\em views}) for each sample. In this context, attribute outliers correspond to point outliers, while class outliers are similar to contextual outliers in the measure that are samples labelled differently across {\em views}.


Existing methods for detection of outliers can be categorized into global and local approaches. Global methods are population based and they model the distribution in the feature space of a set of (annotated) samples. Population distribution can be modelled using either parametric global descriptors or unsupervised clustering approaches. These methods are suited to detect point (or attribute) outliers. Local methods are based on a description of the structure of each sample's neighbors in the feature space and, thus, they are better suited to detect, both, attribute and class outliers.

In this work we introduce  a local approach, which takes benefit of topological properties of neighboring samples,  to detect class and attribute outliers in both single view and multiview data. 

\section{Related work}
\label{sec:SoA}


\subsection{Global methods}

Global approaches seek to estimate parametric data distributions and define outliers as points not folowing the estimated distribution. For instance, \cite{Yang:2009} use Gaussian mixture models (GMM) for outlier detection in which they define each point as a cluster and the outlierness score is the reciprocal of the point likelihood. In general, data distribution cannot be properly estimated  in small sample size sets since  outliers  become influential points  who deviates  global approaches from normal population. This leads to  lack of reproducibility and drop their potential for outlier detection. 

With the advent of big data and deep learning techniques this main drawback of global approaches seems to be mitigated since they can learn complex data structures from big amount of data.  A main issue when dealing with big data is labelling enough samples for training and testing deep learning methods, which is especially difficult in such an imbalanced classification task. \cite{Chalapathy:2019} define four groups of outlier detectors methods: unsupervised, semi-supervised, hybrid and One-Class neural networks. While is clear which methods belong to unsupervised and semi-supervised methods, hybrid methods use deep learning architectures for feature extraction and then use traditional outlier detectors. Finally, One-Class neural network (OC-NN) method is inspired by kernel methods on one-class classification tasks~\citep{Chalapathy2018a}.  A variant of OC-NN architectures is  Deep Support Vector Data Description (Deep-SVDD)~\citep{ruff2018deep}. In that work, the authors train a deep neural network to extract common variation factors  by mapping close inlier data instances to the center of a hyper-sphere. Generative adversarial active learning (GAAL) networks have also been used for outlier detection~\citep{Liu:2018}. 

There are some real use cases in which there are not enough data for learning end-to-end methods. Although domain adaptation and transfer learning techniques can be used to deal with small datasets, the success of these end-to-end methods relies on their capability to learn specific features for the outlier detection task. However, there are some real use case scenarios, like clinical decision support systems or personalized models, in which feature vector are defined beforehand for the particular task and hence cannot be modified, or changed, to detect outliers.

\subsection{Local methods}

Local methods are based on a description of each sample's neighborhood usually defined using the Euclidean distance among samples in the feature space. Given that local methods define outliers in terms of such distances, they are distribution free and, thus, better suited for unbalanced small datasets.

Most local approaches, like \cite{Ramaswamy:2000} or \cite{Angiulli:2002}, define outliers in terms of the distance to the $k$-th nearest point. LOF is an outlier detector method defined in the context of knowledge discovery in databases that assigns an outlierness score to each sample based on local information~\citep{Breunig:2000}. This score is computed in terms of the distance to the $k$ nearest neighbors of each point, meaning a score near to 1 to not be an outlier while higher values provides higher certainty of being it. LOCI~\citep{Papadimitriou:2002} also bases on $k$-th nearest neighbours to define a multi-granularity deviation factor (MDEF) as outlierness measure. The MDEF is the relative deviation of sample's local neighborhood density from the average local neighborhood density, so that a point is an outlier if its MDEF is sufficient large. This way LOCI is effective to detect point outliers and collective outliers, as well. The Isolation Forest, IF, technique~\citep{Liu:2012} builds a tree that isolates {\em attribute} outliers using a binary search. Since an attribute outlier has different values compared to inlier points IF detects them as points such that the length path to reach them is significantly shorter than the mean length to reach any other point.

The selection of the parameters defining neighborhoods is a main bottleneck in local approaches. In particular, the selection of the number $k$ of nearest neighbors is crucial since it greatly affects the performance of methods. Therefore, several strategies for optimal selection of the parameter $k$ have been proposed since the early years of nearest neighborhood approaches. 

\subsection{Graph based methods}


Neighboring relationship can be defined in terms of distances but also in terms of {\em friendship} relationship on structured data, like graphs. In graph structured data and networks, outliers are also linked to topological variation of subgraphs that broke a repetitive pattern. To detect these local structural singularities, it has also been proposed outlier detectors for graph-based representations. 

The early \cite{Brito1997} studied the relationship between connectivity of mutual-knn graph (MKG) and outlier detection, providing a criteria in terms of the graph topology. In particular, they studied the geometric properties of the underlying data points distribution and derived a theoretical criteria to set a value of $k$ ensuring that the connected components of the graph correspond to clusters in the feature space. In that context, outliers were those samples which did not belong to any of this connected components, that is, they correspond to single node connected components. The work in \cite{Brito1997} was later extended in~\cite{Maier2007}. There, the authors provide further insights on the mutual k-nn graphs to derive tighter bounds to estimate the optimal $k$ to build a MKG. A main inconvenience for a practical use is that these bounds are still hard to compute with real data having class outliers.

 \cite{Ning:2018PRL_mutual_k} propose an algorithm to search the optimal $k$ to build the MKG. In that work, the authors introduce the concepts of {\em stability state} and {\em appropriate $k$ (apk)}  for a MKG and they propose and algorithm to search the optimal $k$. Finally, the most recent work of \cite{Wang:2019} also proposes a variation of MKG to minimize the impact of $k$. In their approach, they compute multiple local proximity graphs for $k$ sampled uniformly in a range of values. Then, their approach does not rely in finding the {\em optimal} $k$ of the MKG but in combining the information of all MKGss using a random walk to detect outliers. 

Aside fixing the parameter $k$, another concern about existing local approaches is that outlier scores are defined from the structure of a single neighborhood defined using distances. From a mathematical point of view, this implies that the topological structure of feature spaces is modelled as a norm or metric space~\citep{Munkres}. Although Euclidean spaces admit a topology defined from a metric or norm, these approaches might fail to properly describe more complex spaces (like manifolds ~\citep{Munkres}). In this context, topology is a powerful mathematical approach to model the structure of complex manifolds without the assumption of any parametric model for the data. 

Methods for the detection of communities in social networks can provide a mean to extract a set of topological neighbours from MKG. In this context, attribute outliers are often non connected, or hardly connected, individuals. Meanwhile, class outliers correspond to community members with a user profile, or interests, far of most of community members and they are often ignored. While detecting attribute outliers is done using graph topological properties, computation of a topology in non-structured spaces given by a discrete set of population samples still remains a challenge for topology. \cite{Deb2017}  presented a local method based on neighborhoods given by the communities of the graph built from the samples distances. Despite the promising results in the diagnosis of lung cancer in confocal images, this topology given by graph communities is prone to exclude many points that could not be considered isolated attribute outliers~\citep{Mielgo}. Besides, like other local methods, a main concern is the impact of the parameters used to compute the graph used to detect communities. In case $k$ is too small, communities might exclude points that are not are actually outliers~\citep{Mielgo}, while increasing $k$ produces a single community including all points. Thus, the method requires a proper accurate value for the parameter $k$, which the authors fine-tuned to give optimal results.

Direct analysis of the local structural properties of graphs and networks also allows the detection of outliers and anomalies. The OddBall method  is a widely used method for anomaly detection in networks which focus on detecting nodes having topological properties significantly different compared to neighboring nodes~\citep{Akoglu:2010}. More recently, \cite{Elliot:2019} propose an extension of the NetEMD network method~\citep{wegner:2018} to detect graph anomalies and spectral localization statistics in financial transaction networks. Other methods base on the the Minimum Description Length (MDL) principle. In \cite{Noble:2003} anomalous sub-graphs are detected using variants of MDL. \cite{Eberle:2007} use MDL as well as other probabilistic measures to detect several types of graph anomalies (e.g. unexpected/missing nodes/edges). 



\subsection{Multiview methods}

All methods described above are specifically designed for the detection of attribute outliers in single view problems. Multimodal, or multiview, data can benefit of the different sources of information to detect data outliers. In particular, class outliers are easily spotted when a sample is labelled differently across views. In this context, \cite{HOADGao13} detect class and attribute outliers based on the spectral analysis of the combined adjacency matrix. In that paper, those samples that lie in the kernel space of the combined adjacency matrix are identified as outliers. A different approach is the one proposed in~\cite{Zhao2015} and \cite{Zhao2018}. In that works, the authors propose a generalized K-means method that learns cluster label consistencies across views. Samples having different cluster labels are classified as class outliers. A limitation specific to multi-view methods is the combination of information across views, which usually leads to under-detection of abnormalities arising in single views. Finally, another limitation is that being designed for more than view, multi-view methods are prone to perform poorly in single view problems.

\subsection{Our Contributions}

\begin{figure}
    \centering
    \input{figs/pipeline.tex}
    \caption{Overview of the method: (a) Two-dimensional feature space  with 2 classes (black dots and red crosses). (b) Graph encoding mutual k-nearest neighbors with nodes colored in red and black according to its class. Nodes 6 and 10 correspond, respectively, to the class attribute and attribute outliers. (c) Detection of communities: initial communities, $\C_1$, $\C_2$, in left graph and their extension in the right graph. (d) Feature space and classifier margin, $C$, giving the final outlierness measure from the communities.}
    \label{fig:pipeline}
\end{figure}
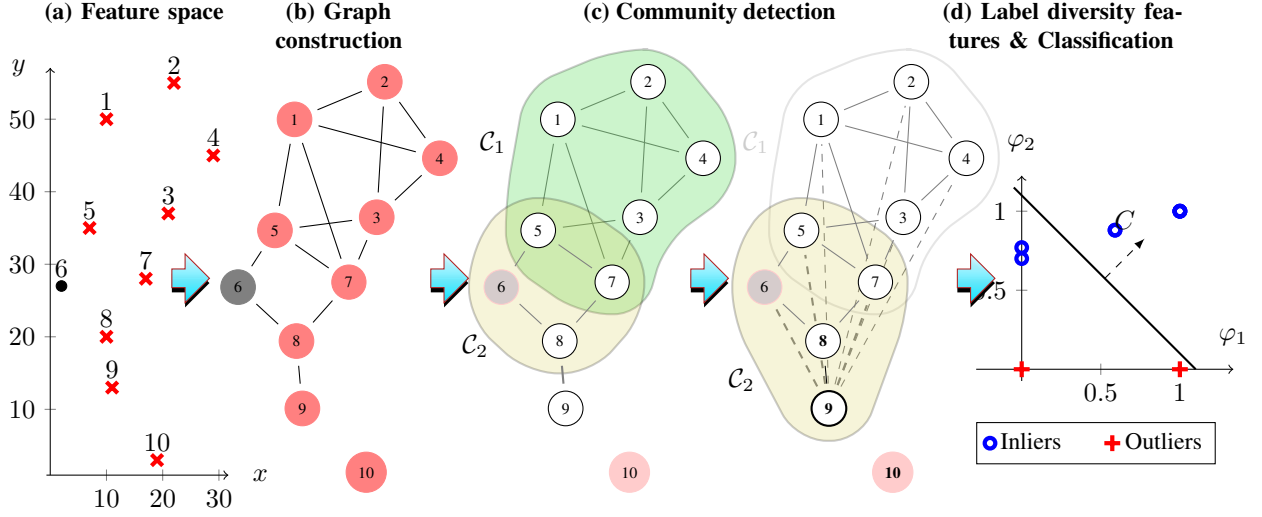

In this work we present a local approach based on a measure of sample labels diversity in a set of topological neighbourhoods of each sample. We compute this topology extending the graph structural method presented in~\cite{Deb2017}. In particular, we refine the initial topology given by community detection methods to include isolated non-outlier points. Sample diversity is computed using probabilistic measures, which summarize the variability of sample labels in the set of topological neighbourhoods samples belong to. These diversity measures provide a normalized {\em outlierness} representation feature space. This normalized space is independent of the actual sample features and only depends on their topological structure. Finally, a  classifier, like support vector machine or random forest,  is used in a final step to detect outliers in this representation space (see \figurename~\ref{fig:pipeline} for further details). We call our method Community-based Outlier Detector, COD. 




Comparing to existing outliers detection methods, the proposed approach have the following two main contributions:

\begin{itemize}
    \item It is based on a topological description of the structure of feature spaces based on the communities of a MKG.
     
    \item This description provides an outlierness representation space based on the intrinsic topological structure of outliers, which is independent of the original feature space.  
    
    \item It is a {\em training cheap} method. As said above, the source dataset used to train the outlier classifier does not significantly bias the performance of the COD method. This advantage is useful when dealing with small datasets or high dimensionality data, since we can  trained a classifier  on a big labelled dataset and then    use it as an off the shelf classifier.
\end{itemize}

In summary, we propose a cheap outlier detector that uses a standard classifier but is trained on meaning-full feature space.




    
    

\section{COD: Outlier Detection based on a Topological Measure of Pattern Diversity}

Our method is a local approach based on the communities  of a graph encoding the structure of the feature space. \figurename~\ref{fig:pipeline} sketches the main steps of our method for the two-dimensional single-view space shown in \figurename~\ref{fig:pipeline}.a. The dataset has 2 classes (black dots and red crosses), one class outlier (numbered 6)  and one attribute  outlier (numbered 10). The 3 main steps of the proposed method applied to the synthetic data  are shown in \figurename~\ref{fig:pipeline}.b-d. First, we encode the local structure of samples using the graph representing their mutual k-nearest neighbor. The nodes of the graph are colored using each class color (red and black). Second, we use methods for dynamical analysis of social networks to compute graph communities  from an initial set of communities. \figurename~\ref{fig:pipeline}.c shows the initial set of communities on the left and their extension on the right image. These extended communities define in the original feature space a set of neighborhoods of each sample. By definition of class and attribute outlier, isolated nodes not belonging to any community are attribute outliers, while class outliers should belong to communities with an heterogeneous distribution of labels. In order to characterize the latter, we define a local measure of abnormality from two probabilistic measures  of each sample heterogeneity computed in its set of neighborhoods. These two measures define a function from the set of samples to the unit square that maps inliers and outliers to different corners of the square. A classifier, $C$, discriminating between inliers and outliers provides our measure of outlierness, \figurename~\ref{fig:pipeline}.d.

In the next sections, we give details about each of the main
steps: graph construction, community detection and definition of
the feature space for the classification of inliers and outliers.

\subsection{Graph Construction}

The graph, $G=(D,A)$,  is given by the adjacency matrix of the mutual k-nearest neighbor of the set of samples. Let $D:=\{(\V^i,\ell_{\V^i}) \hspace{0.25cm} | \hspace{0.25cm} \V^i=(v_1^i,\dots,v_n^i)\in \R^n, \hspace{0.25cm} \ell_{\V^i} \in \{1, \dots, n_l\}\}_{i=1}^N$ be a set of $N$ labelled points in an n-dimensional feature space endowed with a distance, namely $d$. For any positive integer, $k$ , let $\kNN(\V^i)$ denote the set of $\V^i$ k-nearest neighbors and $\mkNN(V^i)$ the set of $\V^i$ mutual k-nearest neighbors defined as: %
\begin{equation}
\mkNN(\V^i):=\{\V^j \mbox{ such that } \V^j \in \kNN(\V^i) \mbox{
and } \V^i \in \kNN(\V^j) \}
\end{equation}

\noindent Then, the  adjacency matrix, $A=(a(\V^i,\V^j))_{ij}=(a_{ij})_{ij}$, codifying $G$ is defined as: %
\begin{equation}\label{eq:adjacency_weights}
a_{ij}=\left \{
\begin{array}{lcl}
\frac{1}{d(\V^i,\V^j)+1} & \mbox{if} & \V^j \in \mkNN(\V^i)\\
0 & & \mbox{otherwise}
\end{array}
\right .
\end{equation}
\noindent for $d(\V^i,\V^j)$ the distance between $\V^j$ and $\V^i$. We note that the graph edges, $a_{ij}$, are  in $[0,1]$, being close to 0 if mutual neighbors are far from each other, close to 1, otherwise. This way, the number of neighbors for every node, $\V^i$, is related to the sparseness of the point $\V^i$ in the feature space.

\subsection{Community Detection}

In order to alleviate the impact of the parameters involved in the computation of \eqref{eq:adjacency_weights}, communities are computed using criteria for dynamic computation of communities ~\citep{iLCD,DynComDetect11,DynCommDetect16} to extend an initial set of communities. The initial communities are given by Percolation clusters \citep{Percolation} which are defined as
maximal unions of adjacent k-cliques. 
Percolation communities are prone to exclude many points that are not actual attribute outliers~\citep{Mielgo}. In order to add them to the set of initial communities, we extend them following a modification of the community detector proposed in~\cite{iLCD}. An isolated node, $\W^j$, is added to
a initially detected  community, $\C$, if it fulfills that: %
\begin{equation}\label{Eq:AddCrit}
\CS(\C,\W^j) \geq \delta \IC(\C)
\end{equation} %
\noindent for $\delta\in [0,1]$ a tolerance parameter, $\IC(\C)$ a measure of the community internal connectivity and $\CS(\C,\W^j)$ a measure of the connectivity between $\W^j$ and the community $\C$. Both measures are computed from a function of the degree of the community nodes as follows.

Let $\S$ be the set composed of all  nodes that belong to any of the initially detected communities and  $G^{\S}$ and  $G^{\C}$ the subgraphs induced by $\S$ and $\C$, respectively. Then, for $\forall \V^i \in \C$ we can define the following function, $\rho_{\C}(\V^i)$, measuring its belongingness to the community: %
%
%
\begin{equation}\label{Eq:RhoC}
\rho_{\C}(\V^i):=\frac{\deg^{G^{\C}}(\V^i)
}{\deg^{G^{\S}}(\V^i) }
\end{equation}
\noindent being $\deg$ the  degree function of a node  in a graph. The measure of the internal connectivity of  $\C$ is defined from $\rho_{\C}(\V^i)$ as: %
%
%
\begin{equation}\label{Eq:IC}
\IC(\C):=\sum_{\V_i \in \C} \rho_{\C}(\V_i)
\end{equation}

The measure of the connectivity between $\W^j$ and $\C$ is also defined from $\rho_{\C}(\V^i)$ as: %
\begin{equation}\label{Eq:CS}
\CS(\C,\W^j):=\sum_{\V^i \in \C} \frac{ \rho_{\C}(\V^i)}{d(\W^j,\V^i)+1}
\end{equation}

\noindent since $\CS(\C,\W^j)$ is a weighted average of $\rho_{\C}(\V^i)$ with weights $\frac{1}{d(\W^j,\V^i)+1}$, we have that: %
\begin{equation}
\CS(\C,\W^j) \geq \left ( \max_{ji} \frac{1}{d(\W^j,\V^i)+1} \right ) \IC(\C)
\end{equation}
By the above inequality, we could set $\delta=\max_{ji}
\frac{1}{d(\W^j,\V^i)+1}$. However, such extreme value could
aggregate some attribute outliers to the initial set of
communities. In order to avoid such an artifact, we propose to
define $\delta$ as a percentile of $\frac{1}{d(\W^j,\V^i)+1}$
probabilistic distribution.

\subsection{Outlierness Feature Space}

Given that communities define a set of neighbors in the feature space,
nodes not belonging to any of the extended communities correspond
to attribute outliers. Meanwhile, class outliers are expected to
belong to communities with either high heterogeneity in nodes
label or the majority of nodes with a label different from the
class outlier label. 

Under the above considerations, we define two quantities,
$\varphi_1$, $\varphi_2$, measuring how homogeneous the labels of
the communities a sample belongs to are. For each sample, the function
$\varphi_1$ quantifies the heterogeneity in community labels, while
$\varphi_2$ quantifies how many nodes in the community have a label
different from the sample label. Both measures are based on the
probabilistic distribution of the community node labels and are
normalized in $[0,1]$ in such a way that they define a function
$\varphi$:
\begin{equation}\label{Eq:OutlierNessSingle}
\begin{array}{rccl}
\varphi: & D & \longrightarrow &[0,1] \times [0,1] \\
 & (\V^i,\ell_{\V^i}) & \mapsto & (\varphi_1(\V^i,\ell_{\V^i}), \varphi_2(\V^i,\ell_{\V^i}))
\end{array}
\end{equation}
\noindent mapping inliers to $(1,1)$, attribute outliers to
$(0,0)$ and class outliers to either $(1,0)$ or $(0,1)$ depending
on whether they belong to one or more communities. In case $\V$ does not belong to any community, we have an attribute outlier and, thus, we set $\varphi(\V)=(0,0)$.
This way,
$\varphi$ defines a 2-dimensional feature space able to discriminate inliers and
outliers. The probability of a classifier trained to discriminate
between them provides our outlierness score and its output class
our outlier detection. 

The function $\varphi_1$ measuring the heterogeneity in community labels is computed from their entropy as follows. For each sample $\V$, let $\S_{\V}$ denote the set of communities containing $\V$ and $Ent_{\C}$ the entropy of the labels of the nodes in a community $\C \in \S_{\V}$ defined as: %
\begin{equation}\label{Eq:Entc}
Ent_{\C} =- \sum_{i=1}^{n_l} p_i^{\C} \log(p_i^{\C})
\end{equation}

\noindent for $p_i^{\C}$ the probability in $\C$ of the $i$-th label. This probability is approximated by the proportion of $\C$ nodes that are labelled $i$: %
\begin{equation}\label{Eq:pi}
p_i^{\C}:= \frac{| \{\W \in \C \hspace{0.25cm} | \hspace{0.25cm}
\ell_{\W}=i\}|}{ | \{\W  \in \C \}|}
\end{equation}

\noindent for  $| \cdot |$ denoting the cardinality of a set. Then, $\varphi_1$ is defined as: %
\begin{equation}\label{Eq:Rho1}
\varphi_1(\V,\ell_{\V}):= \varphi_1(\V):= \frac{1}{ | \S_{\V} |} \sum_{\C \in \S_\V}  \pi(Ent_{\C} \leq T )  
\end{equation} %

\noindent being $T \in [0,1/n_{\ell}]$ a tolerance threshold on the maximum entropy allowed in community labels and $\pi(a)=1$ if $a$ holds. 
The score $\varphi_1(\V) \in [0,1]$ has extreme values $\varphi_1(\V)=0$ in case all communities in $\S_{\V}$ have heterogeneous labels and $\varphi_1(\V)=1$ in case the label within each community is the same for all nodes in the community. We note that, in this last case, by definition of $Ent_{\C}$, the label must be the same for all communities in $\S_{\V}$.

The function $\varphi_2$ measuring  how many nodes in the community have a label equal to the sample label, $\V$, is computed from the probability of $\ell_{\V}$, as follows. For each $\C \in \S_{\V}$, let $p_{\ell_{\V}}^{\C}$ denote the probability of $\ell_{\V}$ in $\C$ excluding $\V$: %
\begin{equation}
p_{\ell_{\V}}^{\C}=\frac{| \{ \W \in \C\setminus \V
\hspace{0.25cm} | \hspace{0.25cm} \ell_{\W}=\ell_{\V} \} |}{| \{
\W \in \C\setminus \V \}|}
\end{equation} %
Then, $\varphi_2$ is defined as: %
\begin{equation}\label{Eq:Rho2}
\varphi_2(\V,\ell_{\V}) :=\frac{1 }{ |\S_{\V} |} \sum_{\C \in \S_{\V}} p_{\ell_{\V}}^{\C}
\end{equation}

The measure $\varphi_2(\V,\ell_{\V}) \in [0,1]$ and has extreme values $\varphi_2(\V,\ell_{\V})=0$ in case no community in $\S_{\V}$ is consistent with $\V$ label, $\varphi_2(\V,\ell_{\V})=1$ in case all communities have nodes with labels equal to $\ell_{\V}$. 

Finally, in order to increase the separability across the different types of outliers, we transform $\varphi_1$, $\varphi_2$ to logarithmic scale by applying the function: %
\begin{equation}
    f(x) = \frac{1}{1-\log x}
\end{equation}
\noindent to $x=\varphi_i$, $i=1,2$.

In the multi-view case, we model the space as a Cartesian product and, thus, compute a mutual k-nearest neighbor graph for each view. Communities are computed independently for each of these graphs and so are the probabilistic measures. These view-dependent measures are aggregated to define the function that maps samples to the unit square. We compute the two measures for each view and aggregate them to define $\varphi$. If $\varphi_1^j$, $\varphi_2^j$ denote the measures computed for the $j$-th view, then their minimum values across views defines the function $\varphi$ as: %
\begin{equation}\label{Eq:OutlierNess}
\begin{array}{rccl}
\varphi: & D & \longrightarrow &[0,1] \times [0,1] \\
 & (\V^i,\ell_{\V^i}) & \mapsto & (\varphi_1(\V^i,\ell_{\V^i}), \varphi_2(\V^i,\ell_{\V^i}))
 :=(\min_j \varphi_1^j(\V^i,\ell_{\V^i}), \min_j
 \varphi_2^j(\V^i,\ell_{\V^i}))
\end{array}
\end{equation}

\section{Experiments}

\subsection{Experimental Set-up}

The performance of the proposed method has been assessed in UCI
\footnote{https://archive.ics.uci.edu/ml/datasets.php} datasets
altered to have different $\%$ of attribute and class outliers in,
both, single and multi-view settings. Table \ref{Tab:UCISets}
reports the main characteristics of the UCI datasets selected for
these experiments. The single-view problem was defined directly
using UCI data sets features. Following \cite{HOADGao13}, the
multi-view case was defined by splitting UCI features into
disjoint sets, each of them defining one view. We considered 2 and
3 views for all sets, with the exception of Iris, which
dimensionality only allows splitting features in 2-views.

\begin{table}[ht]
    \centering
    \begin{tabular}{c|ccc}
        Datasets &  Dimension & Num of Samples & Num of Classes \\ \hline
        Iris  & 4 & 150 & 3 \\
        BCW      & 10  & 699 & 2 \\
        Ionosphere          & 34 & 351 & 2 \\
        Letter recognition       & 16 & 20,000 & 26 \\
    \end{tabular}
    \caption{Selected UCI Datasets}
    \label{Tab:UCISets}
\end{table}

We have followed the experimental settings described in
\cite{Zhao2018}. In particular, we have considered 3 combinations
of percentages in attribute and class outliers: 1) 8\% class
outlier  + 2\% attribute outlier, labelled 8-2, 2) 5\%
class outlier  + 5\% attribute outlier, labelled 5-2 and 3) 2\% class outlier  + 8\% attribute outlier, labelled 2-8. To
simulate attribute outliers, the features of the selected samples
were changed by random numbers following a distribution with the
highest probability outside the range of the values expected for
the original data. In the multi-view case, features were altered
in each view. To simulate class outliers, we swapped the labels of
points randomly selected from random pairs of classes. In the
multi-view case, classes were swapped in views randomly selected.

For each outlier configuration, we repeated the experiment 50 times for statistical analysis of results. Our proposal has been compared to state-of-art single and multi view methods methods in terms of Area Under the ROC Curve (AUC). Single-view methods include 4 local and 3 global approaches. The local methods are LOF \citep{Breunig:2000}, LOCI \citep{Papadimitriou:2002}, KNN \citep{Ramaswamy:2000} and IF \citep{Liu:2012}, while the global ones are GMM \citep{Yang:2009}, the graph-based APS \citep{Wang19} and the deep-learning one SO-GAAL \citep{Liu:2018} model. The multi-view methods are the best performers reported in \citep{Zhao2018}, DMOD \citep{Zhao2018}, AP \citep{APAlvarez13} and MLRA \citep{Li15}, and the pioneering multi view approach HOAD \citep{HOADGao13}.

Our method was computed using the following parameters. The
$\mkNN$ graph was computed setting $k=40$. For the extension of
the initial communities, the tolerance $\delta$ in
(\ref{Eq:AddCrit}) was computed for the $75\%$ percentile.

Finally, for the computation of the outlierness measure, we use a SVM classifier trained on a sub-set of the MirFlickr dataset \citep{huiskes08} altered to have the different types of outliers. To trained it,  we use the second last layer of a pre-trained VGG network as feature vector  on images with a single annotation. Then, we applied a K-means to identify those annotations that does not highly overlap between them and, to further reduce the size of samples, we select only those samples with minimal overlapping.  Then, we follow \cite{Zhao2018} to generate class and attribute outliers.




\subsection{Results}

Table \ref{Table:ResultsSingleView} reports the AUC (average $\pm$
standard deviation) for the results obtained for the proposed COD,
LOF, LOCI, IF, GMM, SO-GAAL and APS  on Iris, Breast Cancer Winconsin (BCW),
Ionosphere and Letter Recognition with different settings \citep{Wang19}. Following \citep{Zhao2018}, the first and second best performers for each configuration are marked in red and blue, respectively. The analysis of the ranges lead to the following observations.

According to the number of times methods are top 2, best performers are COD ($16/24=67\%$ times), KNN ($14/24=58\%$ times), IF ($4/24=17\%$ times) and GMM ($7/24=30\%$ times). The remaining methods achieve top ranges in less than $25\%$ of the cases, being the deep-learning method SO-GAAL the worst performer with $0$ top ranges. The ranking of top ranges indicates that local methods perform better than global ones, as 3 best ranked methods are local approaches. It is worth noticing that our COD is the only method that has top ranges for all datasets and outlier configurations. Meanwhile, the other top ranked methods achieve their best performance only in some of the datasets. In particular, LOCI did not converge for the Letter dataset. 

If we compare ranges between configurations with same quantity of attribute outliers but different quantity of class outliers (2-8 against 0-8, 5-5 against 0-5 and 8-2 against 0-2) we have that, in general, methods are better detecting attribute outliers. In particular, the configuration that has worst ranges is the one with higher number of class-outliers (8-2). Among all methods, our COD is the one that has the lowest drop in performance in the presence of class-outliers. 

\begin{table}
 \caption{AUC values (mean $\pm$ standard deviation) for the Single View Case
 }\label{Table:ResultsSingleView}
 \begingroup
\setlength{\tabcolsep}{3pt}
\begin{tabular}{cccccccc}
\hline
 DataSet & Method &   2-8 & 5-5 &  8-2  & 0-8 & 0-5 & 0-2 \\
\hline   & LOF & 0.973 $\pm$ 0.020   & \color{blue}{0.958 $\pm$ 0.024}  & \color{blue}{0.949 $\pm$ 0.039} & \color{red}{0.992 $\pm$ 0.003} & 0.978 $\pm$ 0.003 & 0.958 $\pm$ 0.024  \\
        & LOCI  & 0.962 $\pm$ 0.022 &  0.888 $\pm$ 0.052 &  0.728 $\pm$ 0.058 & 0.971 $\pm$ 0.007  &  0.966 $\pm$ 0.011 &  0.962 $\pm$ 0.006 \\
        & KNN  & \color{blue}{0.974 $\pm$ 0.019}  & 0.936 $\pm$ 0.042 & 0.866 $\pm$ 0.061 & \color{blue}{ 0.990 $\pm$ 0.002} &  \color{blue}{0.982 $\pm$ 0.004} &  \color{blue}{ 0.970 $\pm$ 0.002} \\
Iris    &  IF & 0.905 $\pm$ 0.025  & 0.855 $\pm$ 0.029 &  0.814 $\pm$ 0.036 & 0.987 $\pm$ 0.009  & 0.975 $\pm$ 0.001 & 0.959 $\pm$ 0.000 \\
        &  APS &  0.882 $\pm$ 0.047  & 0.891 $\pm$ 0.046 &  0.882 $\pm$ 0.047 &  0.862 $\pm$ 0.07  &  0.854 $\pm$ 0.088 &  0.910 $\pm$ 0.120 \\
  & GMM  & 0.484 $\pm$ 0.004  & 0.484 $\pm$ 0.004 &  0.485 $\pm$ 0.012 & 0.484 $\pm$ 0.003  &  0.484 $\pm$ 0.003 &  0.484 $\pm$ 0.002 \\
  & SO-GAAL  & 0.614 $\pm$ 0.019 & 0.605 $\pm$ 0.096 & 0.559 $\pm$ 0.087  & 0.663 $\pm$ 0.090  & 0.631 $\pm$ 0.090 &  0.602 $\pm$ 0.134 \\
  & COD  & \color{red}{0.976 $\pm$ 0.045} &  \color{red}{0.980 $\pm$ 0.029} &  \color{red}{0.979 $\pm$  0.020} & 0.971 $\pm$ 0.052 &  \color{red}{0.980 $\pm$ 0.041}& \color{red}{0.989 $\pm$ 0.007}\\

\hline   & LOF  & 0.545 $\pm$ 0.092  &  0.528 $\pm$ 0.071 &  0.509 $\pm$ 0.032 & 0.513 $\pm$ 0.071  &  0.525 $\pm$ 0.088 &  0.511 $\pm$ 0.044 \\
& LOCI  & 0.882 $\pm$ 0.008 &  0.735 $\pm$ 0.010 &  0.593 $\pm$ 0.013 & 0.981 $\pm$ 0.005  &  0.972 $\pm$ 0.004 &  0.963 $\pm$ 0.002 \\
& KNN  & \color{red}{0.889 $\pm$ 0.005}  &  \color{blue}{0.739 $\pm$ 0.008} & \color{blue}{ 0.592 $\pm$ 0.012} & \color{red}{0.991 $\pm$ 0.002}  &  \color{red}{0.981 $\pm$ 0.002} &  \color{blue}{0.966 $\pm$ 0.001} \\
BCW & IF &  \color{blue}{0.885 $\pm$ 0.005} & 0.733 $\pm$ 0.009 &  0.588 $\pm$ 0.013  & \color{blue}{0.987 $\pm$ 0.001}  &  \color{blue}{0.973 $\pm$ 0.001} &  0.959 $\pm$ 0.000 \\
  & APS  &  0.794 $\pm$ 0.029  &  0.674 $\pm$ 0.029 &  0.580 $\pm$ 0.029 & 0.882 $\pm$ 0.033  &  0.863 $\pm$ 0.040 &  0.895 $\pm$ 0.048  \\
  
  & GMM  & 0.371 $\pm$ 0.008  &  0.472 $\pm$ 0.015 &  0.574 $\pm$ 0.015 & 0.299 $\pm$ 0.003  &  0.299 $\pm$ 0.002 &  0.298 $\pm$ 0.002 \\
  & SO-GAAL  & 0.693 $\pm$ 0.040  &  0.597 $\pm$ 0.032 &  0.533 $\pm$ 0.023 & 0.733 $\pm$ 0.056  & 0.709 $\pm$ 0.063 &  0.672 $\pm$ 0.084 \\
  
  & COD &  0.879 $\pm$ 0.039  &  \color{red}{0.830  $\pm$ 0.035} &   \color{red}{0.747 $\pm$ 0.033} & 0.936 $\pm$  0.042  &  0.954 $\pm$ 0.037 & \color{red}{0.978 $\pm$ 0.002}\\

\hline   & LOF &  0.679 $\pm$ 0.034  &  0.647 $\pm$ 0.030 &  0.572 $\pm$ 0.019 & 0.736 $\pm$ 0.041  &  0.797 $\pm$ 0.047 & 0.892 $\pm$ 0.059 \\
& LOCI  & 0.823 $\pm$ 0.025  & 0.722 $\pm$ 0.016 &  0.585 $\pm$ 0.020 & 0.922 $\pm$ 0.027  &  0.956 $\pm$ 0.023 &  0.954 $\pm$ 0.018 \\
& KNN  & \color{blue}{0.874 $\pm$ 0.009}  &  0.734 $\pm$ 0.011 &  0.583 $\pm$ 0.017 & \color{blue}{0.987 $\pm$ 0.005}  &  \color{blue}{0.976 $\pm$ 0.002} &  \color{blue}{0.960 $\pm$ 0.001} \\
Ionosphere & IF &  0.871 $\pm$ 0.011  &  0.734 $\pm$ 0.012 &  0.584 $\pm$ 0.019 & \color{blue}{0.987 $\pm$ 0.006}  & 0.974 $\pm$ 0.001  &  0.958 $\pm$ 0.010  \\
  & APS &  \color{red}{0.881 $\pm$ 0.025}  &  \color{blue}{0.742 $\pm$ 0.026} &  0.579 $\pm$ 0.034 & \color{red}{0.995 $\pm$ 0.002}  &  \color{red}{0.996 $\pm$ 0.004} &  \color{red}{0.996 $\pm$ 0.005} \\
 
 & GMM  & 0.590 $\pm$ 0.012  &  0.697 $\pm$ 0.017 &  \color{red}{0.814 $\pm$ 0.019} & 0.497 $\pm$ 0.001  &  0.497 $\pm$ 0.001 &  0.497 $\pm$ 0.001 \\
  & SO-GAAL  & 0.515 $\pm$ 0.035   &  0.503 $\pm$ 0.029  & 0.511 $\pm$ 0.025  & 0.514 $\pm$ 0.048  & 0.509 $\pm$ 0.043 &  0.503 $\pm$ 0.055 \\
  
  & COD  & 0.845 $\pm$ 0.023 & \color{red}{0.762 $\pm$ 0.032}  & \color{blue}{0.686 $\pm$ 0.033} & 0.902 $\pm$ 0.007  &  0.891 $\pm$ 0.006& 0.890 $\pm$ 0.003 \\
\hline
& LOF & 0.533 $\pm$ 0.009 & 0.522 $\pm$ 0.007 &  0.508 $\pm$ 0.007 & 0.539 $\pm$ 0.011  &  0.527 $\pm$ 0.013 & 0.504 $\pm$ 0.015 \\
& LOCI  & $\pm$  &  $\pm$ &  $\pm$ & $\pm$  &  $\pm$ &  $\pm$ \\
& KNN  & 0.513 $\pm$ 0.010  &  0.500 $\pm$ 0.008 &  0.486 $\pm$ 0.006 & 0.516 $\pm$ 0.012   &  0.510 $\pm$ 0.012 &  0.521 $\pm$ 0.014 \\

Letter Rec. & IF &  0.516 $\pm$ 0.013  &  0.503 $\pm$ 0.010 &  0.494 $\pm$ 0.009  & 0.514 $\pm$ 0.013  &   0.509 $\pm$ 0.016 &  0.534 $\pm$ 0.019 \\
  & APS &  0.492 $\pm$ 0.014 &  0.483 $\pm$ 0.010 & 0.473 $\pm$ 0.011 & 0.503 $\pm$ 0.015  &  0.640 $\pm$ 0.015 & 0.751 $\pm$ 0.014  \\
  
  & GMM  & \color{red}{0.894 $\pm$ 0.010}   & \color{red}{0.893 $\pm$ 0.009}  &  \color{blue}{0.901 $\pm$ 0.008} & \color{red}{0.916 $\pm$ 0.011}  &  \color{red}{0.939 $\pm$ 0.011} &  \color{red}{0.968 $\pm$ 0.012} \\
  & SO-GAAL  & 0.495 $\pm$ 0.021 &  0.492 $\pm$ 0.020 &  0.487 $\pm$ 0.018 & 0.485 $\pm$ 0.024  & 0.493 $\pm$ 0.035 &  0.492 $\pm$ 0.043\\
  & Our COD  &  \color{blue}{0.844 $\pm$ 0.013} & \color{blue}{0.872 $\pm$ 0.009} &  \color{red}{0.906 $\pm$ 0.007} & \color{blue}{0.821 $\pm$ 0.017}  & \color{blue}{0.809 $\pm$ 0.015} & \color{blue}{0.795 $\pm$ 0.013} \\

\hline
\end{tabular}
\endgroup
\end{table}


 Table \ref{Table:ResultsMultiView} reports the AUC (average $\pm$
standard deviation) for the results obtained for the proposed COD,
HOAD, AP, MLRA and DMOD on Iris,  Breast Cancer Winconsin (BCW), Ionosphere and Letter Recognition with
different settings. As before, the best two performers for each configuration are
marked in red and blue. The analysis of the ranges lead to the following
observations.

Ranking as before according to the number of top ranges, the best performers by large are COD ($21/21=100\%$ times) and DMOD ($17/21=81\%$ times). The proposed COD is the best performer for all cases, but three cases (2-View Ionosphere with outlier configuration 2-8 and 3-View Letter with outlier configuration 8-2) that it is the second best. The HOAD method is the worst performer with 0 times having top ranges. 

All methods, excluding the proposed COD, perform better in the 2-view case. Our COD is the only method that has similar (top) ranges for, both, 2 and 3 view configurations. Regarding outlier configurations, there is not a clear trend across datasets. For Iris and Breast performance increases with the number of class outliers, while it decreases for Ionosphere and Letter datasets. Although this behaviour holds for both views, the decrease rate seems to be a bit higher for the 3-view case. Therefore, we attribute it to, both, the separability of the original dataset, as well as, the partition of features to simulate the multi-view configuration, which might have selected features having the least discriminating power. 

\begin{table}
 \caption{AUC values (mean $\pm$ standard deviation) for the Multi View Case
 }\label{Table:ResultsMultiView}
  \begingroup
\setlength{\tabcolsep}{3pt}
\begin{tabular}{c@{}cccccccc}
\hline
 &  & \multicolumn{3}{c}{2-View Case} & &\multicolumn{3}{c}{3-View Case}\\

\cline{3-5} \cline{7-9} DataSet & Method &  2-8 & 5-5 & 8-2 & & 2-8 & 5-5 & 8-2 \\
\hline   & HOAD & 0.167 $\pm$ 0.057 &   0.309 $\pm$ 0.063&  0.430 $\pm$ 0.055 & & $--$ & $--$ & $--$ \\
& AP & 0.326 $ \pm$ 0.027 & 0.630 $ \pm$ 0.021 & \color{blue}{0.840 $ \pm$0.021} & & $--$ & $--$ & $--$ \\
Iris & MLRA & 0.856 $ \pm$ 0.063 & 0.828 $ \pm$ 0.080 & 0.826 $ \pm$ 0.089 & & $--$ & $--$ & $--$ \\
 & DMOD & \color{blue}{0.909 $\pm$ 0.044}  &  \color{blue}{ 0.831 $\pm$ 0.038}&  0.799 $\pm$ 0.068 & & $--$ & $--$ & $--$ \\
  & COD  & \color{red}{0.975 $\pm$ 0.024} &  \color{red}{0.971 $\pm$ 0.023} &  \color{red}{0.970 $\pm$ 0.021} & & $--$ & $--$ & $--$\\

\hline   & HOAD  & 0.555 $\pm$ 0.072 & 0.586  $\pm$ 0.061 & 0.634 $\pm$ 0.046 & & 0.538 $\pm$0.027  &  0.597 $\pm$ 0.038 &  0.643 $\pm$ 0.008\\
& AP & 0.293 $ \pm$ 0.012 & 0.532 $ \pm$ 0.024 & \color{blue}{0.693 $ \pm$
0.023} & & 0.190 $ \pm$ 0.016 & 0.388 $ \pm$ 0.012 & 0.593 $ \pm$ 0.046 \\
BCW & MLRA & 0.745 $ \pm$ 0.056 & 0.715 $ \pm$ 0.022 & 0.688 $ \pm$ 0.028 & & 0.614 $ \pm$ 0.057 & 0.596 $ \pm$ 0.032 & 0.599 $ \pm$ 0.029 \\
  & DMOD  & \color{blue}{0.824 $\pm$ 0.022} & \color{blue}{0.752 $\pm$ 0.019} & 0.692 $\pm$ 0.036 & & \color{blue}{0.657 $\pm$ 0.017} &   \color{blue}{0.720 $\pm$ 0.013} &  \color{blue}{0.799 $\pm$ 0.016} \\
  & COD & \color{red}{0.890 $\pm$ 0.027} & \color{red}{0.935 $\pm$ 0.019} & \color{red}{0.947 $\pm$ 0.013} & &
  \color{red}{0.838 $\pm$ 0.022} &  \color{red}{0.897 $\pm$ 0.020} &  \color{red}{0.910 $\pm$  0.014}\\

\hline   & HOAD & 0.446 $\pm$ 0.074 & 0.442 $\pm$ 0.051 & 0.448 $\pm$ 0.041 & & 0.489 $\pm$ 0.079 &   0.477 $\pm$ 0.072 &  0.444 $\pm$ 0.065\\
& AP & 0.623 $ \pm$ 0.033 & 0.761 $ \pm$ 0.025 & \color{red}{0.822 $ \pm$ 0.030} & & 0.511 $ \pm$ 0.027 & 0.659 $ \pm$ 0.043  & 0.758 $ \pm$ 0.035\\
Ionosphere & MLRA & 0.645$ \pm$ 0.084 & 0.669 $ \pm$  0.028& 0.776$ \pm$ 0.037 & & 0.645 $ \pm$ 0.040 & 0.663 $ \pm$ 0.048& 0.700 $ \pm$ 0.045\\
  & DMOD & \color{red}{0.877 $\pm$ 0.032} & \color{blue}{0.801 $\pm$ 0.042} & 0.774 $\pm$ 0.049 & & \color{blue}{0.818 $\pm$ 0.018}  & \color{blue}{0.787  $\pm$ 0.039} &  \color{blue}{0.784 $\pm$ 0.037}\\
  & COD  & \color{blue}{0.841 $\pm$ 0.024} & \color{red}{0.811 $\pm$ 0.024} & \color{blue}{0.780 $\pm$ 0.029} & & \color{red}{ 0.854 $\pm$  0.019 }&   \color{red}{0.827 $\pm$  0.025}&  \color{red}{0.791 $\pm$  0.036} \\
\hline

& HOAD &  0.536 $\pm$ 0.046 & 0.663 $\pm$ 0.057 & 0.569 $\pm$ 0.049 & & 0.193 $\pm$ 0.022 & 0.488 $\pm$ 0.111& 0.563 $\pm$ 0.081\\
& AP & 0.372 $ \pm$ 0.057& 0.550 $ \pm$ 0.043 & 0.640 $ \pm$ 0.051 & & 0.189 $ \pm$ 0.039 & 0.340 $ \pm$ 0.037 & 0.570 $ \pm$ 0.63\\
Letter Rec. & MLRA & 0.883 $ \pm$ 0.024 & 0.817$ \pm$ 0.051 & \color{blue}{0.786 $ \pm$ 0.065} & & 0.841 $ \pm$ 0.055 & 0.716 $ \pm$ 0.044 & 0.640 $ \pm$ 0.081 \\
  & DMOD & \color{blue}{0.912 $\pm$ 0.029} & \color{blue}{0.846 $\pm$ 0.022} &  0.762 $\pm$ 0.025 & & \color{red}{0.916 $\pm$ 0.031}  & \color{blue}{0.815 $\pm$ 0.038}&  \color{blue}{0.664 $\pm$ 0.037} \\
  & COD  & \color{red}{ 0.926 $\pm$ 0.009} & \color{red}{0.904 $\pm$ 0.011} &  \color{red}{0.877 $\pm$ 0.011} & & \color{blue}{0.843 $\pm$ 0.014} & \color{red}{0.816 $\pm$ 0.016} & \color{red}{0.774 $\pm$ 0.017 }\\

\hline
\end{tabular}
\endgroup
\end{table}


\subsection{Discussion}

The method has been tested on 4 UCI datasets altered to have different configurations of class and attribute outliers in single and multi view spaces. Datasets have been selected to  be representative of the main configurations of classification feature spaces. In particular, they include datasets presenting the most common artifacts dropping performance of methods, like small sample size (Ionosphere), large dimensionality (Ionosphere) and large number of classes with some of them being minority groups (Letter Recognition). The method has been compared to several state of the art methods for detection of outliers in single and multi view settings. The analysis of results lead to the following conclusions.

In, both, single and multi view settings, local methods perform better than global ones. In fact, the worst performers in both settings are the global methods SO-GAAL (single view) and HOAD (multi view). Surprisingly, the deep learning approach SO-GAAL is the worst performer in the single view setting without any range in the top two. Although deep learning approaches achieve excellent results in classification problems, they require huge amounts of data to model population distribution. Being based in big data, their capability to model rare cases (like outliers) might be, as our experiments indicate, limited compared to other approaches.

Among all local approaches, the proposed COD outperforms existing methods in, both, single and multi view settings, regardless of the outlier configuration. Unlike most methods, it performs equally well in small size and high dimensionality datasets. It is worth noticing that COD has been applied with the same parameter configuration to all datasets in, both, training and test. Regarding COD training, we used a completely different repository (MirFLick) and no learning transfer was applied. This is a main advantage compared to existing approaches that require fine tuning of parameters and a re-training for new datasets.

Although the choice of parameters (the $k$ for the computation of the mutual k-nearest neighbor graph, in particular) is not critical, the performance of COD significantly increases for low dimensional balanced datasets like Iris. We think that the combination of all information obtained from graphs computed sampling the range of possible values of $k$ could improve the performance of COD in high dimensional unbalanced datasets. 

\section{Conclusion}

This paper presents a local approach for outlier detection based on a outlierness representation space that codifies samples label diversity using a topological description of samples structure in feature space. This representation space is normalized and describes the intrinsic structural properties of feature spaces independently of the particular dataset. It follows that COD needs only to be trained once and can be applied to any data set without any transfer learning. 

The reported experiments show that COD outperforms  existing methods in, both, single and multi view settings, regardless of the outlier configuration. Although the choice of parameters  is not critical, the performance of COD significantly increases for low dimensional balanced datasets like Iris. We think that the description of the feature space topology could be more elaborated and include higher order aspects (like closed paths and cycles). Algebraic topology is a unique mathematical discipline that provides with very low-level descriptions of complex manifolds in high dimensional spaces without the need of either exhaustive training or access to big data. In particular, the groups of persistent homology provide an algebraic description of the structure of point clouds. In order to define intrinsic properties, these groups are computed using a set of neighbourhoods (called filtration) which increases according to a parameter. Parameter values are sampled to obtain such set of neighbourhoods and those properties that are more stable across the filtration constitute the persistent homology. Our current efforts focus on the improvement of COD incorporating the computation of persistent homology to better define space topological properties.

\section*{Acknowledgment}
This work was supported by Spanish projects RTI2018-095645-B-C21 and RTI2018-095209-B-C21,
Generalitat de Catalunya, 2017-SGR-1624 and 2017-SGR-1783 and CERCA-Programme and the ATTRACT project funded by the EC under Grant Agreement 777222. 
Debora Gil is supported by Serra Hunter Fellow. The Titan X Pascal and Titan V used for this research was donated by the NVIDIA Corporation. DGil is a Serra Hunter Fellow.

\bibliographystyle{unsrtnat}
\bibliography{references} 
\bibliography{referencesDeb}

\end{document}

%% file: figs/pipeline.tex
    \begin{tikzpicture}[node distance=1.5cm,
    attstyle/.style={draw,fill=gray!40,minimum size=15,drop shadow},
    entitystyle/.style={shape=circle,draw,minimum size=15,fill=white,drop shadow},
    arrowAttstyle/.style={dashed,gray},
    arrowRelstyle/.style={},
    style1/.style={
      matrix of math nodes,
      every node/.append style={text width=#1,align=center,minimum height=2mm},
      nodes in empty cells,
      left delimiter=[,
      right delimiter=],
      },
    mypin/.style={draw=gray, shorten <=1mm, shorten >=1mm},
    nod/.style={draw,circle,fill=white,font=\tiny, inner sep=1mm, outer sep=1mm}
  ]


    

    \begin{axis}[anchor=east,width=4cm,height=7cm,
    ylabel=$y$,
    xlabel=$x$,
    axis lines=middle,
    axis line style={->},
    x label style={at={(current axis.right of origin)},right=2mm},
    y label style={at={(current axis.above origin)},anchor=north, left=2mm},
    enlarge x limits={abs=2mm},
    enlarge y limits={abs=2mm},
    ]

    \addplot[visualization depends on={\thisrow{id}\as\myvalue},
        scatter/classes={%
    A={mark=x,draw=red,mark size=3pt,ultra thick},
    B={draw=black}
    },
    scatter,only marks,scatter src=explicit symbolic,
    nodes near coords*={\pgfmathprintnumber[int detect]\myvalue}] table[x=x,y=y,meta=class]  {data/points.txt};
    \node[outer sep=1mm] at (axis cs:10,50) (c) {};
    \node[outer sep=1mm] at (axis cs:15,0) (o) {};
    \end{axis}
    
    
   \node[align=center,anchor=north,text width=3cm]  at ($ (o) + (0,65mm) $) (lp) {\bf \footnotesize (a) Feature space};

    \node[nod,draw=red!50,fill=red!50] at ($ (c) + (25mm,0) $) (node_g_1) { 1};
    \node[nod,draw=red!50,fill=red!50] at ($(node_g_1) +(12mm,5mm)$) (node_g_2) { 2};
    \node[nod,draw=red!50,fill=red!50] at ($(node_g_1) +(-50:17mm)$) (node_g_3) { 3};
    \node[nod,draw=red!50,fill=red!50] at ($(node_g_1) +(-15:20mm)$) (node_g_4) { 4};
    \node[nod,draw=red!50,fill=red!50] at ($(node_g_1) +(-100:15mm)$) (node_g_5) {5};
    \node[nod,draw=black!50,fill=black!50] at ($(node_g_5) +(-123:9mm)$) (node_g_6) {6};
    \node[nod,draw=red!50,fill=red!50] at ($(node_g_5) +(-35:12mm)$) (node_g_7) {7};
    \node[nod,draw=red!50,fill=red!50] at ($(node_g_5) +(-79:15mm)$) (node_g_8) {8};
    \node[nod,draw=red!50,fill=red!50] at ($(node_g_8) +(-85:9mm)$) (node_g_9) {9};
    \node[nod,draw=red!50,fill=red!50] at ($(node_g_9) +(-45:12mm)$) (node_g_10) {10}; 

    \draw (node_g_1) -- (node_g_2)  -- (node_g_3)  -- (node_g_4);
    \draw   (node_g_1) -- (node_g_5) -- (node_g_3);
    \draw   (node_g_1) -- (node_g_4) -- (node_g_2);
     \draw   (node_g_3) -- (node_g_7);
     \draw   (node_g_1) -- (node_g_7);
                            
    \draw (node_g_5) -- (node_g_6)  -- (node_g_8)  -- (node_g_7) -- (node_g_5);
    
    \draw   (node_g_8) -- (node_g_9); 

    \node[rounded corners=5mm,fit=(node_g_1) (node_g_2) (node_g_3) (node_g_4) (node_g_5) (node_g_6) (node_g_7) (node_g_8) (node_g_9) (node_g_10),inner sep=0mm,outer sep=-2mm] (gc) {};

    \node[text width=30mm,align=center,anchor=north]  at  (lp.north-| gc)  (lgc) {\bf \footnotesize (b) Graph construction};

        
    \node[nod] at ($(node_g_1) + (35mm,0)$) (node_cd_1) {1};
    \node[nod] at ($(node_cd_1) +(12mm,5mm)$) (node_cd_2) {2};
    \node[nod] at ($(node_cd_1) +(-50:17mm)$) (node_cd_3) {3};
    \node[nod] at ($(node_cd_1) +(-15:20mm)$) (node_cd_4) {4};
    \node[nod] at ($(node_cd_1) +(-100:15mm)$) (node_cd_5) {5};
    \node[nod,draw=red!20,fill=red!20!black!20] at ($(node_cd_5) +(-123:9mm)$) (node_cd_6) {6};
    \node[nod] at ($(node_cd_5) +(-35:12mm)$) (node_cd_7) {7};
    \node[nod] at ($(node_cd_5) +(-79:15mm)$) (node_cd_8) {8};
    \node[nod] at ($(node_cd_8) +(-85:9mm)$) (node_cd_9) {9};
    \node[nod,draw=red!20,fill=red!20] at ($(node_cd_9) +(-45:12mm)$) (node_cd_10) {10}; 

    \begin{scope}[on background layer] 
    \draw[draw=gray] (node_cd_1) -- (node_cd_2)    -- (node_cd_3)  -- (node_cd_4); 
    \draw[draw=gray]   (node_cd_1) -- (node_cd_5) -- (node_cd_3);
    \draw[draw=gray]   (node_cd_1) -- (node_cd_4) -- (node_cd_2);
     \draw[draw=gray]   (node_cd_3) -- (node_cd_7);
     \draw[draw=gray]   (node_cd_1) -- (node_cd_7);
     \end{scope}
    
    \begin{scope}[on background layer] 
    \draw[rounded corners=4mm, thick, fill=green!80!black,opacity=0.2]
    ($(node_cd_2.north) + (0,1mm)$) --    ($(node_cd_2.east  |- node_cd_2.north) + (0,1mm)$)  -- ($(node_cd_4.north-|node_cd_4.east) + (1mm,1mm)$)   --
     ($(node_cd_4.south-|node_cd_4.east) + (1mm,-1mm)$)   -- ($(node_cd_3.south-|node_cd_3.east)  + (0mm,0mm)$)  -- 
     ($(node_cd_7.south-|node_cd_7.east)  + (1mm,-1mm)$)  -- ($(node_cd_7.south-|node_cd_7.west)  + (-1mm,-1mm)$)  -- 
     ($(node_cd_5.west |-node_cd_5.south) + (-1mm,-1mm)$) --  ($(node_cd_5.west |-node_cd_5.north) + (-1mm,1mm)$) -- node[left,midway,opacity=1] {$\C_1$}
     ($(node_cd_1.north -| node_cd_1.west)+ (-1mm,1mm)$) -- ($(node_cd_2.west  |- node_cd_2.north) + (-1mm,1mm)$)  -- ($(node_cd_2.north) + (0,1mm)$)  ;
     \end{scope}
                            
    \draw[draw=gray] (node_cd_5) -- (node_cd_6)  -- (node_cd_8)  -- (node_cd_7) -- (node_cd_5);
    
    \begin{scope}[on background layer]
    \draw[rounded corners=4mm, thick, fill=yellow!80!black,opacity=0.2]
    ($(node_cd_5.north)+ (-0mm,1mm)$) -- ($(node_cd_5.north -| node_cd_5.east)+ (1mm,1mm)$) -- 
    ($(node_cd_7.north-|node_cd_7.east) + (1mm,1mm)$)  -- ($(node_cd_7.south-|node_cd_7.east) + (1mm,-1mm)$)  -- 
    ($(node_cd_8.south -| node_cd_8.east) + (1mm,-1mm)$)   -- ($(node_cd_8.south -| node_cd_8.west) + (-1mm,-1mm)$)   -- node[left,midway,opacity=1] {$\C_2$}
    ($(node_cd_6.south -| node_cd_6.west)  + (-1mm,-1mm)$)   --  ($(node_cd_6.north -| node_cd_6.west)  + (-1mm,1mm)$)   -- 
    ($(node_cd_5.north -| node_cd_5.west)+ (-1mm,1mm)$) -- ($(node_cd_5.north) + (0mm,1mm)$);
    \end{scope}
    \draw[draw=gray,thick]   (node_cd_8) -- (node_cd_9);


    \node[rounded corners=5mm,fit=(node_cd_1) (node_cd_2) (node_cd_3) (node_cd_4) (node_cd_5) (node_cd_6) (node_cd_7) (node_cd_8) (node_cd_9) (node_cd_10),inner sep=0mm,minimum height=5.5cm,outer sep=-2mm] (cd) {};

    \node at (gc.east) (a1) {\tikzfancyarrow[5mm]{}};
    
    
        
    \node[nod] at ($(node_cd_1) + (35mm,0)$) (node_gcd_1) {1};
    \node[nod] at ($(node_gcd_1) +(12mm,5mm)$) (node_gcd_2) {2};
    \node[nod] at ($(node_gcd_1) +(-50:17mm)$) (node_gcd_3) {3};
    \node[nod] at ($(node_gcd_1) +(-15:20mm)$) (node_gcd_4) {4};
    \node[nod] at ($(node_gcd_1) +(-100:15mm)$) (node_gcd_5) {5};
    \node[nod,draw=red!20,fill=red!20!black!20] at ($(node_gcd_5) +(-123:9mm)$) (node_gcd_6) {6};
    \node[nod] at ($(node_gcd_5) +(-35:12mm)$) (node_gcd_7) {7};
    \node[nod] at ($(node_gcd_5) +(-79:15mm)$) (node_gcd_8) {\bf 8};
    \node[nod,thick] at ($(node_gcd_8) +(-85:9mm)$) (node_gcd_9) {\bf 9};
    \node[nod,draw=red!20,fill=red!20] at ($(node_gcd_9) +(-45:12mm)$) (node_gcd_10) {\bf 10}; 

     \begin{scope}[on background layer] 
    \draw[dashed,gray,thick] (node_gcd_9) -- (node_gcd_7);
    \draw[dashed,gray,thick] (node_gcd_9) -- (node_gcd_6);
    \draw[dashed,gray,thick] (node_gcd_9) -- (node_gcd_5);
    \draw[dashed,gray] (node_gcd_9) -- (node_gcd_4);
    \draw[dashed,gray] (node_gcd_9) -- (node_gcd_3);
    \draw[dashed,gray] (node_gcd_9) -- (node_gcd_2);
    \draw[dashed,gray] (node_gcd_9) -- (node_gcd_1);
    
    \end{scope}

    \draw[draw=gray] (node_gcd_1) -- (node_gcd_2)  -- (node_gcd_3)  -- (node_gcd_4);
    \draw[draw=gray]   (node_gcd_1) -- (node_gcd_5) -- (node_gcd_3);
    \draw[draw=gray]   (node_gcd_1) -- (node_gcd_4) -- (node_gcd_2);
     \draw[draw=gray]   (node_gcd_3) -- (node_gcd_7);
     \draw[draw=gray]   (node_gcd_1) -- (node_gcd_7);
    
    \begin{scope}[on background layer] 
    \draw[draw=gray, rounded corners=4mm, thick,opacity=0.2]
    ($(node_gcd_2.north) + (0,1mm)$) --    ($(node_gcd_2.east  |- node_gcd_2.north) + (0,1mm)$)  -- ($(node_gcd_4.north-|node_gcd_4.east) + (1mm,1mm)$)   --
     ($(node_gcd_4.south-|node_gcd_4.east) + (1mm,-1mm)$)   -- ($(node_gcd_3.south-|node_gcd_3.east)  + (0mm,0mm)$)  -- 
     ($(node_gcd_7.south-|node_gcd_7.east)  + (1mm,-1mm)$)  -- ($(node_gcd_7.south-|node_gcd_7.west)  + (-1mm,-1mm)$)  -- 
     ($(node_gcd_5.west |-node_gcd_5.south) + (-1mm,-1mm)$) --  ($(node_gcd_5.west |-node_gcd_5.north) + (-1mm,1mm)$) --  node[left,midway] {$\C_1$}
     ($(node_gcd_1.north -| node_gcd_1.west)+ (-1mm,1mm)$) -- ($(node_gcd_2.west  |- node_gcd_2.north) + (-1mm,1mm)$)  -- ($(node_gcd_2.north) + (0,1mm)$)  ;
     \end{scope}
                            
    \draw[draw=gray] (node_gcd_5) -- (node_gcd_6)  -- (node_gcd_8)  -- (node_gcd_7) -- (node_gcd_5);
    
    \begin{scope}[on background layer]
    \draw[rounded corners=4mm, thick, fill=yellow!80!black,opacity=0.2]
     ($(node_gcd_5.north)+ (-0mm,1mm)$) -- ($(node_gcd_5.north -| node_gcd_5.east)+ (1mm,1mm)$) -- 
    ($(node_gcd_7.north-|node_gcd_7.east) + (1mm,1mm)$)  -- ($(node_gcd_7.south-|node_gcd_7.east) + (1mm,-1mm)$)  --  
    ($(node_gcd_9.south -| node_gcd_9.east) + (1mm,-1mm)$)   -- ($(node_gcd_9.south -| node_gcd_9.west) + (-1mm,-1mm)$)   -- node[left,midway,opacity=1] {$\C_2$}
    ($(node_gcd_6.south -| node_gcd_6.west)  + (-1mm,-1mm)$)   --  ($(node_gcd_6.north -| node_gcd_6.west)  + (-1mm,1mm)$)   -- 
    ($(node_gcd_5.north -| node_gcd_5.west)+ (-1mm,1mm)$) -- ($(node_gcd_5.north) + (0mm,1mm)$); \end{scope} 
    
    \draw[draw]   (node_gcd_8) -- (node_gcd_9);


    \node[rounded corners=5mm,fit=(node_gcd_1) (node_gcd_2) (node_gcd_3) (node_gcd_4) (node_gcd_5) (node_gcd_6) (node_gcd_7) (node_gcd_8) (node_gcd_9) (node_gcd_10),inner sep=0mm,minimum height=5.5cm,outer sep=-2mm] (gcd) {};
    
    \node at ($ (cd.west)!.45!(gcd.east) $) (a3) {\tikzfancyarrow[5mm]{}};
    
    \node[align=center,anchor=north]  at (lp.north  -| a3) (lgcd) {\bf \footnotesize (c) Community detection};

        %
    \begin{axis}[at={($(gcd.east)+(0mm,0)$)},anchor=west,width=50mm,
    axis equal,
    scatter/classes={%
    0={mark=o,draw=blue,mark size=2pt,ultra thick},
    1={mark=+,draw=red,mark size=3pt,ultra thick }
    },
    enlarge y limits={value=.1,upper},
    ylabel=$\varphi_2$,
    xlabel=$\varphi_1$,
    axis lines=middle,
    axis line style={->},
    x label style={at={(current axis.right of origin)},above=2mm},
    y label style={at={(current axis.above origin)},anchor=north, above=2mm},
    enlarge x limits={abs=0.5cm},
    legend style={
    at={(0.0,-0.2)},
    anchor=north west,
    legend columns=-1,
    font=\footnotesize,
    /tikz/every even column/.append style={column sep=0.5cm}
    },
    ] 
    
    
    \addplot[scatter,only marks,scatter src=explicit symbolic] table[x=phi1,y=phi2,meta=outlier]  {data/points.txt};
    \legend{Inliers, Outliers}
    \addplot[samples=2,  domain=-0.05:1.1,thick] {1.1 - x }
    coordinate[pos=.5] (a) coordinate (b);
    \node[outer sep=1mm] at (axis cs:0.25,0) (origin) {};
    \end{axis}
    
    \draw[-latex,dashed] (a) -- ($(a)!.75cm!90:(b)$) node[anchor=south east] {$C$};

    \node[align=center,anchor=north,text width=6cm]  at (lgcd.north east -| origin) (lfc) {\bf \footnotesize (d) Label diversity features \& Classification};

     \node at ($ (gc.west) +(-5mm,0) $)  {\tikzfancyarrow[5mm]{}};
     \node at (gcd.east)  {\tikzfancyarrow[5mm]{}};

    \end{tikzpicture}